\documentclass[10pt,journal,compsoc]{IEEEtran}
\usepackage{xcolor}
\usepackage{amssymb}
\usepackage{amsmath}

\ifCLASSOPTIONcompsoc
  \usepackage[nocompress]{cite}
\else
  \usepackage{cite}
\fi

\ifCLASSINFOpdf
  \usepackage[pdftex]{graphicx}
  \graphicspath{{../graph/}}
  \DeclareGraphicsExtensions{.pdf,.jpeg,.png}
\else
\fi

\usepackage{amsmath}
\begin{document}
\title{Relationship-Embedded Representation Learning for Grounding Referring Expressions
}

\author{Sibei~Yang,
        Guanbin~Li,  \emph{Member, IEEE}, and 
        Yizhou~Yu, \emph{Fellow, IEEE}


\thanks{This work was partially supported by the National Key Research and Development Program of China under Grant No. 2019YFC0118100, the National Natural Science Foundation of China under Grant No.61976250 and No.61702565, the Science and Technology Program of Guangdong Province under Grant No.~2017B010116001 and the Hong Kong PhD Fellowship. This paper was presented at the IEEE Conference CVPR, 2019~\cite{yang2019cross-modal}. (Corresponding authors: Guanbin Li and Yizhou Yu).}
\thanks{S. Yang and Y. Yu are with the Department
of Computer Science, The University of Hong Kong, Hong Kong, and Y. Yu is also with Deepwise AI Lab (e-mail: sbyang9@hku.hk; yizhouy@acm.org).}
\thanks{G. Li is with the school of Data and Computer Science, Sun Yat-sen University, Guangzhou 510006, China (e-mail: liguanbin@mail.sysu.edu.cn).}}

%


\IEEEtitleabstractindextext{%
\begin{abstract}
Grounding referring expressions in images aims to locate the object instance in an image described by a referring expression. It involves a joint understanding of natural language and image content, and is essential for a range of visual tasks related to human-computer interaction. As a language-to-vision matching task, the core of this problem is to not only extract all the necessary information (i.e., objects and the relationships among them) in both the image and referring expression, but also make full use of context information to align cross-modal semantic concepts in the extracted information. Unfortunately, existing work on grounding referring expressions fails to accurately extract multi-order relationships from the referring expression and associate them with the objects and their related contexts in the image. In this paper, we propose a Cross-Modal Relationship Extractor (CMRE) to adaptively highlight objects and relationships (spatial and semantic relations) related to the given expression with a cross-modal attention mechanism, and represent the extracted information as a language-guided visual relation graph. In addition, we propose a Gated Graph Convolutional Network (GGCN) to compute multimodal semantic contexts by fusing information from different modes and propagating multimodal information in the structured relation graph. Experimental results on three common benchmark datasets show that our Cross-Modal Relationship Inference Network, which consists of CMRE and GGCN, significantly surpasses all existing state-of-the-art methods.
\end{abstract}

}

\maketitle
\IEEEdisplaynontitleabstractindextext
\IEEEpeerreviewmaketitle

\IEEEraisesectionheading{\section{Introduction}\label{sec:introduction}}
\IEEEPARstart{A} fundamental capability of AI for bridging humans and machines in the physical world is comprehending natural language utterances and their relationship with visual information. This capability is required by many challenging tasks, among which, grounding referring expressions~\cite{kazemzadeh2014referitgame, mao2016generation} is an essential one. The task of grounding referring expressions needs to locate a target visual object in an image by understanding multimodal semantic concepts as well as relationships between referring natural language expressions (e.g.~``the man with sun glasses'', ``the dog near a white car'') and the image content.

Identifying the object proposal referred to by an expression from a set of proposals in an image is a typical formulation of grounding referring expressions~\cite{yu2018mattnet}. Recent methods adopt the combination of Convolutional Neural Networks (CNN)~\cite{krizhevsky2012imagenet} and Long Short-Term Memory Neural Networks (LSTM)~\cite{hochreiter1997long} to process multimodal information in images and referring expressions. CNNs extract visual features of single objects, global visual contexts~\cite{mao2016generation, rohrbach2016grounding} and pairwise visual differences~\cite{Liu_2017_ICCV, yu2018mattnet, yu2016modeling, yu2017joint} while LSTMs encode global language contexts~\cite{Liu_2017_ICCV, luo2017comprehension, mao2016generation, yu2016modeling, yu2017joint} and language features of decomposed phrases~\cite{hu2017modeling, yu2018mattnet, zhang2018grounding}. In addition, the cooperation between CNNs and LSTMs captures the context of object pairs~\cite{hu2017modeling, nagaraja2016modeling, zhang2018grounding}. However, existing work cannot extract all the required information (i.e. individual objects; first-order relationships or multi-order relationships) accurately from referring expressions and the captured contexts in such work also have discrepancies with the contexts described by referring expressions. In this paper, we refer to \cite{zhang2018grounding} and define the ``context'' as the objects as well as their attributes and relationships mentioned in the expression that help distinguish a referred object from other objects.

\begin{figure}[t]
\begin{center}
\includegraphics[width=1\linewidth]{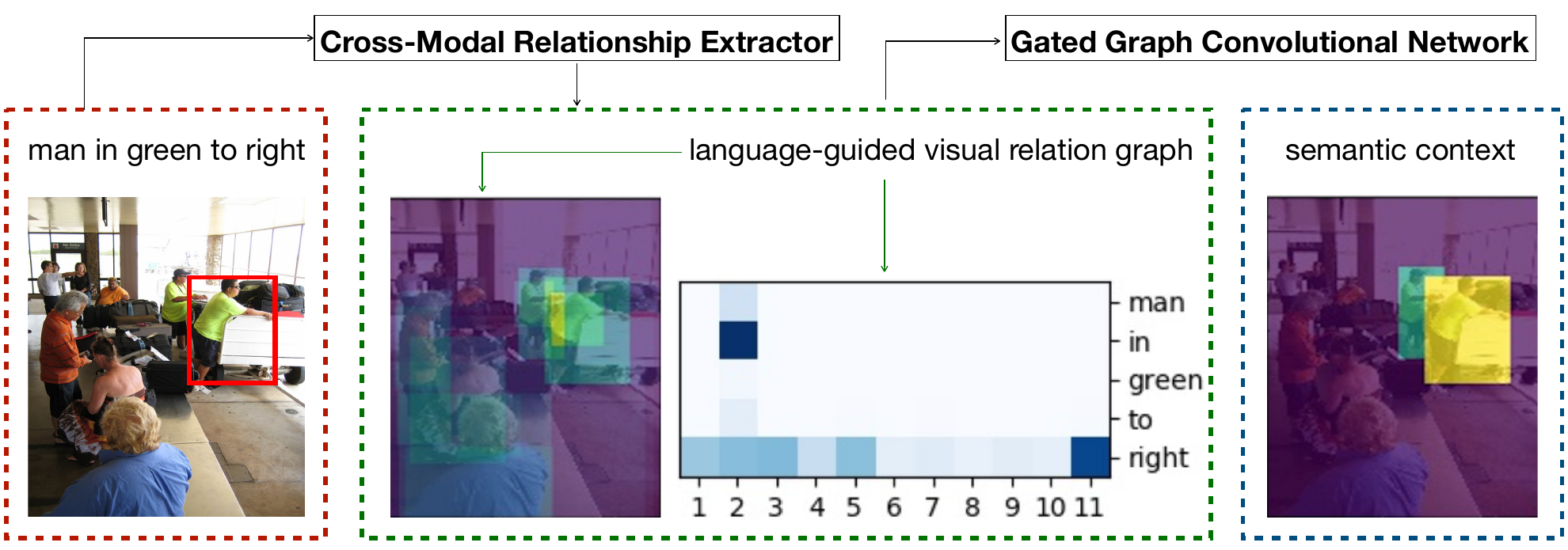}
\end{center}
   \caption{Cross-Modal Relationship Inference Network. Given an expression and image, Cross-Modal Relationship Extractor constructs the language-guided visual relation graphs ({\color{black}spatial relation graph as an example, }the attention scores of proposals and edges' types are visualized inside green dashed box). The Gated Graph Convolutional Network capture semantic context and computes the matching scores between context of proposals and context of expression (the matching scores of proposals are shown inside blue dashed box). Warmer color indicates higher scores of pixels and darker blue indicates higher scores of edges' types.}
\label{fig:intro}
\end{figure}

To solve the problem of grounding referring expressions, the accurate extraction of all required information (i.e. objects and the relationships among them in the image and referring expressions) is crucial for any given pair of expression and image. Because of the unpredictability and flexibility of an expression describing the scene in an image~\cite{mao2016generation}, the proposed model needs to extract the information adaptively. For example, if ``The man holding a red balloon'' is located in an image with two or more men, the nouns/noun phrases (``man'' and ``red balloon'') and the relation word ``holding'' need to be extracted from the natural language expression; meanwhile, proposals for ``man'' and ``red balloon'' and the visual relationship (`holding'') linking them together should be identified in the image. ``The parking meter on the left of the man holding a red balloon'' is a more complicated example, which involves an additional object ``parking meter'' and additional relational information ``left''. In this example, on one hand, there are three individual objects (i.e. ``man'', ``red balloon'' and ``parking meter''), that need to be recognized in both the image and expression. Object proposals in the image can be either obtained with an object detector \cite{ren2015faster} or provided as part of the dataset \cite{yu2016modeling, mao2016generation}. Nouns and noun phrases in the expression need to be extracted, and words in the same phrase should refer to the same object. Unfortunately, existing methods only consider individual words and softly parsed phrases \cite{hu2017modeling, yu2018mattnet, zhang2018grounding}, but words in the same softly parsed phrase cannot be constrained to the same object. On the other hand, the second-order relationship between the target and the ``red balloon'' via the ``man'' need to be {\color{black}inferred from either the detected direct semantic relationship ``holding'' or spatial relationship between object pairs ``on the left of''}. Unfortunately, existing work either does not support relationship modeling or only considers first-order relationships among objects~\cite{hu2017modeling, nagaraja2016modeling, zhang2018grounding}. Theoretically, visual relation detectors~\cite{dai2017detecting, lu2016visual, zellers2018neural} and natural language parsers can help achieve that goal by detecting relational information in the image and parsing grammatical relations among the words in the expression. {\color{black}However, existing visual relation detectors, which focus only on the extraction of semantic relationship, cannot deliver satisfactory  and sufficient clues for highly unrestricted scene compositions~\cite{zhang2018grounding},} and existing language parsers have adverse effects on performance for grounding referring expressions due to their parsing errors~\cite{yu2018mattnet, zhang2018grounding}.

Moreover, the target object is distinguished from other objects on the basis of their contexts and the context of the expression~\cite{nagaraja2016modeling, yu2016modeling, zhang2018grounding}; therefore, accurate and consistent representation of contextual information in the referring expression and object proposals is essential. Nevertheless, existing methods for context modeling either cannot represent the context accurately or cannot achieve high-level consistency between both types of contexts mentioned above, and the reasons are given below. First, noisy information introduced by existing work on global language context modeling~\cite{Liu_2017_ICCV, luo2017comprehension, mao2016generation, yu2016modeling, yu2017joint} and global visual context modeling~\cite{mao2016generation, rohrbach2016grounding} makes it hard to align and match these two types of contexts. Second, pairwise visual differences computed in existing work~\cite{Liu_2017_ICCV, yu2018mattnet, yu2016modeling, yu2017joint} can only represent instance-level visual differences among objects of the same category. {\color{black}Third, existing work on context modeling for object pairs~\cite{hu2017modeling, nagaraja2016modeling, zhang2018grounding} only considers first-order relationships instead of multi-order relationships (e.g., they directly extract the relationship between the pairs of (target, ``man'') and (target, ``balloon'') without considering the  ``man'' is ``holding the balloon'' when extracting the relationship between the target ``parking meter'' and ``the man'').} In addition, multi-order relationships are actually structured information, which cannot be modeled by the context encoders adopted by existing work on grounding referring expressions.

Given the limitations of existing methods, our proposed end-to-end Cross-Modal Relationship Inference Network (CMRIN) aims to overcome the aforementioned difficulties. CMRIN consists of two modules, i.e., the Cross-Modal Relationship Extractor (CMRE) and the Gated Graph Convolutional Network (GGCN). An example is illustrated in Fig.~\ref{fig:intro}. The CMRE extracts all the required information adaptively (i.e., nouns/noun phrases and relationship words from the expressions, and object proposals and their visual relationships from the image) for constructing a language-guided visual relation graph with cross-modal attention. First, CMRE constructs {\color{black}two scene graphs (a spatial relation graph as well as a semantic relation graph)} for the image. Second, it extracts noun phrases in the expression using a constituency tree, meanwhile, it learns to classify the words in the expression into four types and further assign the words/phrases to the vertices and edges in {\color{black}each scene graph}. 
Finally, it constructs the language-guided visual relation graph from the normalized attention distribution of words/phrases over vertices and edges of {\color{black} each scene graph}. The GGCN fuses information from different modes and propagates the fused information in the language-guided visual relation graph to obtain semantic contexts of the expression by performing the following two steps. First, it fuses the contexts in the expression into the visual relation graph to form a multimodal relation graph, which includes the spatial/semantic relationships, visual information and language contexts; Second, gated graph convolutional operations are applied to the multimodal relation graph to obtain the semantic contexts.
We have tested our proposed CMRIN on three common benchmark datasets, including RefCOCO~\cite{yu2016modeling}, RefCOCO+~\cite{yu2016modeling} and RefCOCOg~\cite{mao2016generation}, for grounding referring expressions. Experimental results show that our proposed network outperforms all other state-of-the-art methods.

In summary, this paper has the following contributions:
\begin{itemize}
    \item Cross-Modal Relationship Extractor (CMRE) is proposed to convert the pair of input expression and image into a language-guided visual relation graph. For any given pair of expression and image, CMRE highlights objects as well as {\color{black}spatial and semantic relationships} among them with a cross-modal attention mechanism by considering the words and phrases in the expression as guidance.
    \item Gated Graph Convolutional Network (GGCN) is proposed to capture multimodal semantic context with multi-order relationships. GGCN fuses information from different modes and propagates fused information in the language-guided visual relation graph.
    \item CMRE and GGCN are integrated into Cross-Modal Relationship Inference Network (CMRIN), which outperforms all existing state-of-the-art methods on grounding referring expressions using the ground-truth proposals. In addition, CMRIN shows its robustness using the detected proposals.
\end{itemize}
{\color{black}This paper is an extended version of~\cite{yang2019cross-modal}, it provides a more complete introduction and analysis to the proposed cross-modal relationship inference network for referring expression comprehension, providing additional insights and relevant research discussion, verification of the effectiveness of framework components, network parameter analysis and more elaborated experimental comparisons.} 
Furthermore, we propose to add phrase parsing for the expression and apply it to enhance the representation of language-guided visual relation graphs, which helps to better align linguistic words with visual objects. Second, to complement the spatial relation graph, we have also extracted semantic relations and use them as another guidance in edge gate computation for multi-order relationship inference in our proposed GGCN. Experimental results show that by introducing phrase decomposition for referring expressions and semantic relationship modeling for images, it can bring different levels of performance improvement and make the algorithm more complete and robust.
\section{Related Work}
\subsection{Grounding Referring Expressions}
Grounding referring expression and referring expression generation \cite{mao2016generation} are dual tasks. The latter is to generate an unambiguous text expression for a target object in an image, and the former selects the corresponding object according to the image content referred by a text expression.

To address the problem of grounding referring expression, some previous work \cite{Liu_2017_ICCV, luo2017comprehension, mao2016generation, yu2017joint, yu2016modeling, deng2018visual} extracts visual object features from CNN and treat an expression as a whole to encode language feature through an LSTM. Among them, some methods \cite{luo2017comprehension, mao2016generation, yu2016modeling} learn to maximize the posterior probability of the target object given the expression and the image, and the others~\cite{Liu_2017_ICCV, yu2017joint} model the joint probability of the target object and the expression directly. Specifically, MMI~\cite{mao2016generation} applies the same CNN-LSTM network architecture for grounding referring expressions and referring expression generation respectively, and jointly optimize those two parts together. Speaker~\cite{yu2016modeling} improves MMI by taking more consideration between the comparisons on objects of the same type in the image, and it encodes visual appearance differences and relative spatial (i.e.~location and size) differences between object and surrounding objects of the same object category. Speaker-Listener-Reinforcer~\cite{yu2017joint} proposes an Reinforcer module to sample more discriminative expressions for helping the training of the Speaker. Attr \cite{Liu_2017_ICCV} suggests that the attributes of objects help to distinguish the target object from other ones, and it learns the attributes of objects and encodes the features from the learned attributes and visual features. A-ATT \cite{deng2018visual} adopts joint attention mechanism on query, image and objects multiply round to obtain the communication among the three different types of information. {\color{black}However, all of the above methods independently encode the images and expressions without considering the interactions between them, and the learned monolithic representations in the two modes are not practical to the semantic-rich visual scenes and complex expressions.}

Different from the methods above, Neg Bag \cite{nagaraja2016modeling} proposes to feed the concatenation of visual object representation, visual context representation and the word embedding to an LSTM model. Recent methods \cite{hu2017modeling, yu2018mattnet, zhang2018grounding} learn to decompose an expression into different components and compute the language-vision matching scores of each module for objects. Specially, CMN \cite{hu2017modeling} learns to parse the expression into a fixed form of subject-object-relationship; MAttNet \cite{yu2018mattnet} decomposes the expression into subject, location and relationship modules, and the module weights are computed for combining those three modules. VC \cite{zhang2018grounding} obtains the context-cue language features and referent-cue language features for both single objects and pairwise objects. {\color{black}However, all of the existing works are based on simple expression decomposition and match directly with the detected object features and the additionally computed relationship features~\cite{dai2017detecting, lu2016visual, zellers2018neural}, without considering the cross-modal alignment of multi-order relationship among objects and attributes, they are therefore arduous to adapt to the referring of objects in highly unrestricted scenes. Our Cross-Modal Relationship Extractor also learns to parse the expression, but we treat the parsed words as the guidance to highlight all the objects and their relationships described in the expression automatically to build the language-guided visual relation graphs which are further enhanced by a tailor-designed gated graph neural network for cross-modal multi-order context reasoning and alignment.} 

\subsection{Context Modeling}
Context modeling has been applied in many visual recognition tasks, e.g., object detection \cite{bell2016inside, ren2017accurate,li-crossmodal}, semantic segmentation \cite{chen2018deeplab, zhang2018context} and saliency detection \cite{li-dcl,li-mdf}. For example, ION~\cite{bell2016inside} uses four directional Recurrent Neural Networks (RNNs) to compute the context features on feature maps from four spatial directions. Ren \textit{et al.}~\cite{ren2017accurate} propose Recurrent Rolling Convolution architecture to gradually aggregate context among the feature maps with different resolutions. Context Encoding Module \cite{zhang2018context} encodes the global semantic context by learning an inherent codebook which is a set of visual centers. Recently, Structure Inference Network \cite{liu2018structure} formulates the context modeling task as a graph structure inference problem \cite{jain2016structural, kipf2017semi-supervised, Marino_2017_CVPR}, and it obtains the scene context by applying RNN to proposals in image.

As contextual information helps to distinguish the target from other objects, previous work on grounding referring expressions has also attempted to captured the context. For example, some early works~\cite{mao2016generation, rohrbach2016grounding} propose to encode the entire image as a visual context, but that global contextual information usually cannot accurately match with the local context described by the expression. Other works~\cite{ Liu_2017_ICCV, yu2018mattnet, yu2016modeling, yu2017joint} capture the visual difference between the objects belonging to the same category in an image, but the visual difference of the object's appearance is often insufficient to distinguish the target from other objects. In fact, the visual difference between the context including appearance and relationship is essential, e.g., ``Man holding a balloon'', the necessary information to locate the ``man'' is not only the appearance of the ``man'' but the ``holding'' relation with the ``balloon''. There are also some works~\cite{ hu2017modeling, nagaraja2016modeling, zhang2018grounding} which model the context from the context of object pairs, but they only consider the context with the first-order relationship between the objects. Inspired by Graph Convolutional Network \cite{kipf2017semi-supervised} for classification, our Gated Graph Convolutional Network flexibly capture the context referring to the expression by message passing, and the context with multi-order relationships can be captured.

\begin{figure*}[h]
\begin{center}
\includegraphics[width=1\linewidth]{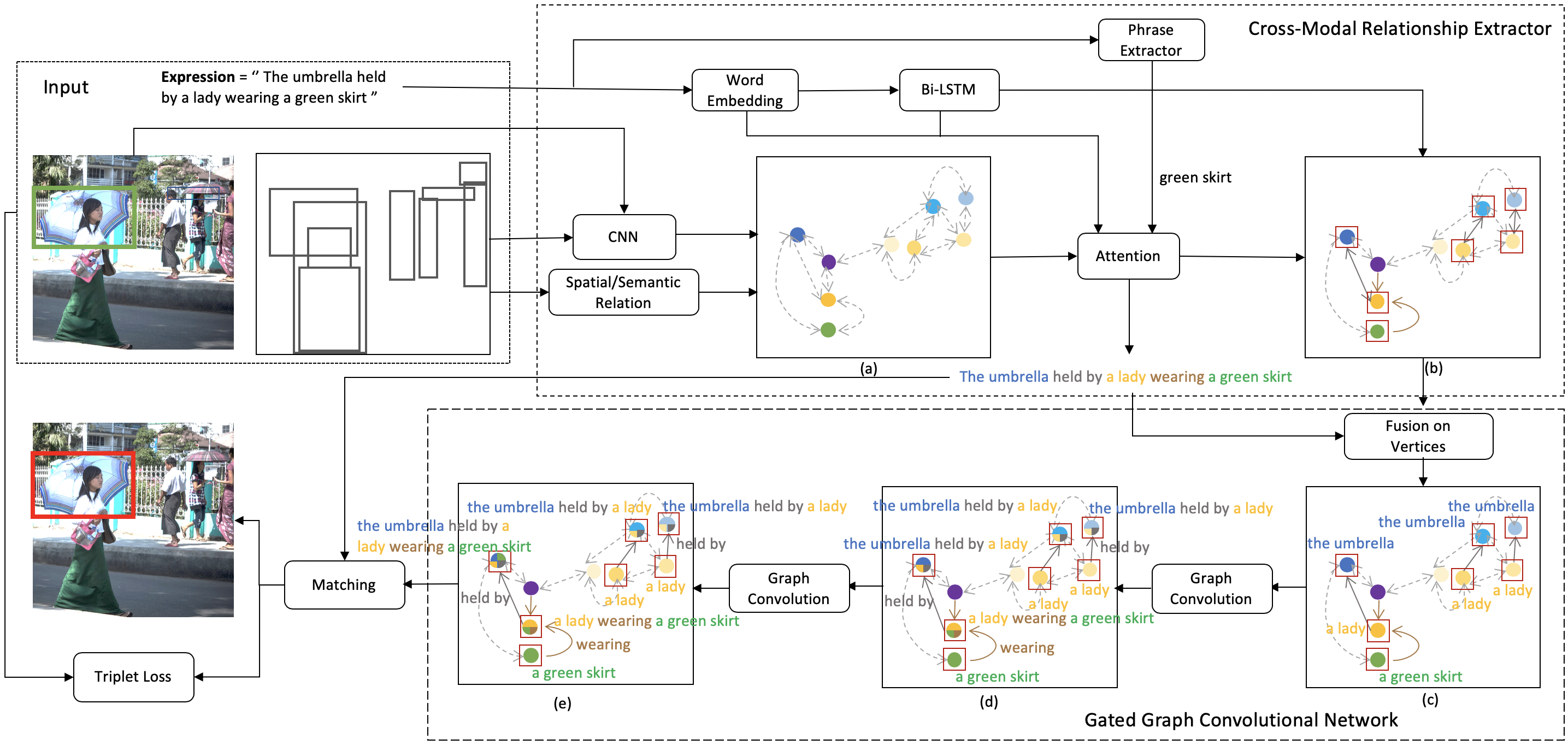}
\end{center}
   \caption{An overview of our Cross-Modal Relationship Inference Network for grounding referring expressions~(better view in color). We use color to represent semantics, i.e. yellow denotes ``person'', green denotes ``green shirt'', blue denotes ``umbrella'', purple means ``white T-shirt'', brown means ``wearing'' and dark grey refers to ``held by''. It includes a Cross-Modal Relationship Extractor (CMRE) and a Gated Graph Convolutional Network (GGCN). First, CMRE constructs (a) a spatial relation graph from the visual features of object proposals and spatial relationships between proposals. Second, CMRE parses the expression into a constituency tree and extracts the valid noun phrases. Third, CMRE highlights the vertices (red bounding boxes) and edges (solid lines) to generate (b) a language-guided visual relation graph using cross-modal attention between words/phrases in the referring expression and the spatial relation graph's vertices and edges. Fourth, GGCN fuses the context of words into the language-guided visual relation graph to obtain (c) a multimodal (language, visual and spatial information) relation graph. Fifth, GGCN captures (d) the multimodal semantic context with first-order relationships by performing gated graph convolutional operations in the relation graph. By performing gated graph convolutional operations multiple iterations, (e) semantic context with multi-order relationships can be computed. Finally, CMRIN calculates the matching scores between semantic context of proposals and the global context of the referring expression. The triplet loss with online hard negative mining is adopted during training and the proposal with the highest matching score is chosen.}
\label{fig:cmrin}
\end{figure*}

\subsection{Vision-Language}
The combination of language and vision has
been extensively studied in the last few years 
due to its significance for building AI systems. Besides grounding referring expressions, image/video captioning~\cite{farhadi2010every,wu2018interpretable} and visual question answering \cite{antol2015vqa:} are two popular and fundamental tasks.

\textbf{Image caption} is to generate image-relevant textual descriptions for given images. Early approaches \cite{farhadi2010every, kulkarni2013babytalk} extract visual concepts (i.e., objects and attributes) from images and format the sentences from those visual concepts and templates. Recently, some work \cite{vinyals2015show, xu2015show} starts to encode the image as visual representations (e.g., single visual representation for the whole image \cite{vinyals2015show} and a set of visual representations for different sub-regions of the image \cite{xu2015show}) by applying CNN, and then decode the visual representations into language descriptions through LSTM. The attention mechanism is adopted to attend the most relevant part of visual information \cite{xu2015show} of it with the already generated text \cite{Lu2017Adaptive} in every time step of LSTM. Some of the recent approaches \cite{ling2017teaching, dai2018a} use the additional information (e.g. phrases and semantic words) extracted from image or text to help generate high quality sentences. {\color{black}There are also some works which focus on the description of a specified object in an image, a.k.a referring expression generation \cite{mao2016generation}, which is a dual problem of visual grounding. It can be applied as an auxiliary component to referring expression comprehension to enhance the performance of cross-modal matching by computing the semantic distance between the generated statement and the given expression \cite{mao2016generation}. However, as the description of the object is varied and involves complex context information and relationships with other objects, the performance improvement for referring expression comprehension of unrestricted complex scenes is limited.} 

\textbf{Visual question answering} is to correctly infer the answer for a given pair of image and textual question. Most of the existing work \cite{ malinowski2015ask, gao2015are, fukui2016multimodal, lu2016hierarchical, schwartz2017high-order, shih2016where, yang2016stacked} extracts the visual features from the image through CNN and encodes the question to language representation by passing the question into LSTM. And then, the answer is predicted by cooperation between those two types of representations. The cooperation is implemented by approaches, like learning a common embedding space for visual and language representations \cite{malinowski2015ask, gao2015are, fukui2016multimodal}, or attending the most discriminate regions of image by applying different attention mechanisms on both representations \cite{lu2016hierarchical, schwartz2017high-order, shih2016where, yang2016stacked}, or using both of them together. Besides the direct prediction of the answers, interpreting the reasoning procedure is important as well. The reasoning procedures are modeled from three different perspectives (i.e. relation-based modeling \cite{santoro2017a, NIPS2018_7311}, attention-based modeling \cite{fukui2016multimodal, yu2017multi-modal} and module-based modeling \cite{andreas2016neural, cao2018visual}). {\color{black}Although visual question answering and referring expression comprehension have different problem definitions and solving goals, visual grounding is the key to endowing VQA with interpretability, which helps to ground their answers to relevant regions in the image~\cite{zhang2019interpretable, cao2018visual}. On the other hand, cross-modal feature fusion and semantics reasoning are equally effective and important for both issues. The study of the two problems can be integrated and learned from each other~\cite{fukui2016multimodal, zhu2016visual7w}.}


\subsection{Graph Neural Networks}
\textcolor{black}{Graph Neural Networks (GNNs) which are widely used to model the relational dependencies among elements of a graph through message passing \cite{battaglia2018relational, kipf2017semi-supervised, velivckovic2017graph}, have been successfully applied to various context-aware visual tasks, e.g., semi-supervised classification \cite{kipf2017semi-supervised}, zero-shot recognition\cite{wang2018zero} and object detection \cite{liu2018structure}.}

\textcolor{black}{Graph-structured representations and GNNs have also been introduced to the tasks of language and vision understanding. The methods in \cite{teney2017graph, norcliffe2018learning, cadene2019murel, yang2019auto} for VQA and image captioning represent an image as a graph structure where the vertices represent visual regions of an image and the edges are relationships among them, and then capture the visual context of each region node by GNN propagation. Specifically, \cite{teney2017graph} and \cite{yang2019auto} encode the contextual features at vertices by using the graph networks based on the recurrent unit \cite{chung2014empirical} and the graph convolutional network (GCN) respectively. Their graph networks operate in the modes of vision and language independently. Different with them, our graph network performs on the top of multi-modal graph to learn the language-guided contexts at vertices. The recent works, \cite{norcliffe2018learning} and \cite{cadene2019murel}, also obtain the convolved graph representations over the language-conditioned graphs: the former identifies neighbors for a vertex as its K most similar vertices and update feature at the vertex as sum of the learned features of its neighbors weighted by the learned weighting factors in each convolution layer, and the latter considers relationships between any pairs of vertices and aggregates the relational features for vertices using max pooling operator. Different with the above methods, we define gates of vertices and edges to implement the different influences of neighbors and relationships, and the gates are learned globally. 
To the best of our knowledge, we are the first to incorporate the graph convolutional networks in referring expressions comprehension for multi-order relationships representation learning.}

\section{Cross-Modal Relationship Inference Network}\label{sec:algorithm}
Our proposed Cross-Modal Relationship Inference Network (CMRIN) relies on relationships among objects and context captured in the multimodal relation graph to choose the target object proposal in the input image referred to by the input expression. First, CMRIN constructs a language-guided visual relation graph using the Cross-Modal Relationship Extractor. Second, it captures multimodal context from the relation graph based on the Gated Graph Convolutional Network. Finally, a matching score is computed for each object proposal according to its multimodal context and the context of the input expression. The overall architecture of our CMRIN for grounding referring expressions is illustrated in Fig.~\ref{fig:cmrin}. In the rest of this section, we elaborate all the modules in this network.
\subsection{Cross-Modal Relationship Extractor}
The Cross-Modal Relationship Extractor (CMRE) adaptively constructs the language-guided visual relation graph according to each given pair of image and expression using a cross-modal attention mechanism. Our CMRE considers both the word level and the phrase level. At the word level, it softly classify the words in the expression into four types (i.e., entity, relation, absolute location, and unnecessary words) according to the context of the words. At the phrase level, it extracts noun phrases, which are directly taken as entity phrases. Meanwhile, the context of the entire expression can be computed from the context of each individual word. In addition, a spatial relation graph of the image is constructed by linking object proposals in the image according to their size and locations {\color{black}and a semantic relation graph is constructed by an off-the-shelf object relationship detector~\cite{yang2018graph}}. Next, CMRE generates the language-guided visual relation graph by highlighting the vertices and edges of the {\color{black}relation graphs}. Highlighting is implemented as computing cross-modal attention between the words/phrases in the expression and the vertices and edges in the {\color{black}relation graphs}.

\subsubsection{\textcolor{black}{Relation Graph Construction}}\label{sec:rgc}
\textcolor{black}{Exploring spatial relations and semantic relations among object proposals within an image is necessary for grounding referring expressions, because they are frequently occurs in referring expressions. Thus, we construct two different graphs by exploring two different types of relationships, i.e., spatial relation graph and semantic relation graph.}

\begin{figure}[h]
\begin{center}
\includegraphics[width=1.0\linewidth]{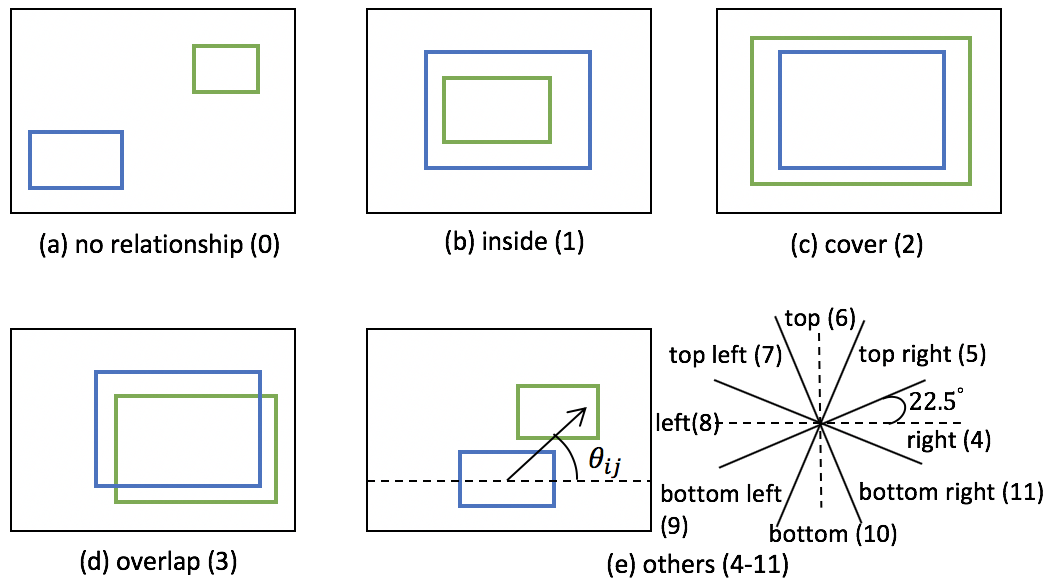}
\end{center}
   \caption{All types of spatial relationships between proposal $o_i$ (green box) and proposal $o_j$ (blue box). The number following the relationship is the label.}
\label{fig:spatial}
\end{figure}

\textcolor{black}{For spatial relation graph, we obtain the spatial relationship between each pair of object proposals according to their size and locations, which bears resemblance to the approach in \cite{yao2018exploring}.} For a given image $I$ with $K$ object proposals (bounding boxes), $O = \{o_i\}_{i=1}^{K}$, the location of each proposal $o_i$ is denoted as $loc_i = (x_i, y_i, w_i, h_i)$, where $(x_i, y_i)$ are the normalized coordinates of the center of proposal $o_i$, and $w_i$ and $h_i$ are the normalized width and height. The spatial feature $\mathbf{p}_i$ is defined as $\mathbf{p}_i = [x_i, y_i, w_i, h_i, w_ih_i]$. For any pair of proposals $o_i$ and $o_j$, the spatial relationship $r_{ij}$ between them is defined as follows. We compute the relative distance $d_{ij}$, relative angle $\theta_{ij}$ (i.e. the angle between the horizontal axis and vector $(x_i-x_j, y_i-y_j)$) and Intersection over Union $u_{ij}$ between them. If $o_i$ includes $o_j$, $r_{ij}$ is set to ``inside''; if $o_i$ is covered by $o_j$, $r_{ij}$ is set to ``cover''; if none of the above two cases is true and $u_{ij}$ is larger than $0.5$, $r_{ij}$ is set to ``overlap''; otherwise, when the ratio between $d_{ij}$ and the diagonal length of the image is larger than $0.5$, $r_{ij}$ is set to ``no relationship''. In the rest of the cases, $r_{ij}$ is assigned to one of the following spatial relationships, ``right'', ``top right'', ``top'', ``top left'', ``left'', ``bottom left'', ``bottom'' and ``bottom right'', according to the relative angle $\theta_{ij}$. The details are shown in Fig.~\ref{fig:spatial}.

The directed spatial relation graph $G^s=(V, E, \mathbf{X}^s)$ is constructed from the set of object proposals $O$ and the set of pairwise relationships $R = \{r_{ij}\}_{i,j=1}^{K}$, where $V = \{v_i\}_{i=1}^{K}$ is the set of vertices and vertex $v_i$ corresponds to proposal $o_i$; $E = \{e_{ij}\}_{i,j=1}^{K}$ is the set of edges and $e_{ij}$ is the index label of relationship $r_{ij}$; $\mathbf{X}^s = \{\mathbf{x}_i^s\}_{i=1}^K$ is the set of features at vertices and $\mathbf{x}^s_i \in \mathbb{R}^{D_x}$ is the visual feature of proposal $o_i$, where $D_x$ is the dimension of visual feature. $\mathbf{x}^s_i$ is extracted using a pretrained CNN model. A valid index label of $E$ ranges from $1$ to $N_e = 11$ (the label of ``no relationship'' is $0$).

\textcolor{black}{Similar to the spatial relation graph $G^s = (V, E, \mathbf{X}^s)$, the semantic relation graph $G^{sem} = (V, \mathbf{E}^{sem}, \mathbf{X}^s)$ shares the same sets of vertices and features at vertices as $G^s$, but instead the set of edges $\mathbf{E}^{sem}$ is extracted by a pretrained object relationship detector \cite{yang2018graph}}.

\textcolor{black}{The spatial relation graph $G^s$ and semantic relation graph $G^{sem}$}, which are constructed from the image, involves the visual features of the proposals as well as their spatial relationships or semantic relationships. They are further transformed into the language-guided visual relation graph based on the guidance from the expression, which will be detailed in Section~\ref{sec:lgvrg}. \textcolor{black}{To simplify the description and focus on the pipeline design of the proposed method, we adopt the $G^s$ as the example for remaining part in Section~\ref{sec:algorithm}. And the detail implementation for the semantic branch will be described in Section~\ref{sec:sg}}.

\subsubsection{Phrase}
Parsing the phrases in the expression is paramount as it helps to accurately highlight the vertices of graph $G^s$ referred to by the expression. For example, if the expression is ``the umbrella held by a lady wearing a green skirt'', it is necessary to recognize the noun phrase (i.e. ``green skirt''), and the words in this phrase refer to the same vertex. We only extract noun phrases in this paper since they are most relevant to the objects in the image. Specifically, 
Our CMRE follows the three steps below to extract the noun phrases. 
First, it parses the given expression into a constituency tree, which breaks the expression into sub-phrases. Second, it locates candidate noun phrases (i.e. ``the umbrella'', ``a lady'' and ``a green skirt'')
from the leaves to the root. On each path from the leaves to the root, it extracts the first noun phrase and ignores the other noun phrases. 
Third, it eliminates determiners and words indicating absolute location in the extracted noun phrase candidates. A candidate phrase is valid if the number of remaining words is at least two. Thus, ``green shirt'' is a valid noun phrase. For a given expression $L = \{l_t\}_{t=1}^T$ ($T$ is the number of words), we denote the set of extracted noun phrases as $Q = \{q_m\}_{m=1}^{M}$, where $M$ is the number of phrases.

\subsubsection{Language Representation}
Inspired by the attention weighted sum of word vectors over different modules in \cite{hu2017modeling, zhang2018grounding, yu2018mattnet}, our CMRE defines attention distributions of words/phrases over the vertices and edges of the spatial relation graph $G^s$. In addition, different words in a referring expression may play different roles. For referring expressions, words can usually be classified into four types (i.e entity, relation, absolute location and unnecessary words), and the type for noun phrases is entity. By parsing the expression into different types of words and distributing words/phrases over the vertices and edges of graph $G^s$, the language embedding of every vertex and edge can be captured, and the global language context can also be obtained.

Given an expression $L = \{l_t\}_{t=1}^T$, CMRE first learns a $D_f$-dimensional embedding for each word, $\mathbf{F}^l = \{\mathbf{f}^l_t \in R^{D_f}\}_{t=1}^{T}$, and then applies a bi-directional LSTM~\cite{schuster1997bidirectional} to encode the context of words. The context of word $l_t$ is the concatenation of its forward and backward hidden vectors, denoted as $\mathbf{h}^l_t \in \mathbb{R}^{D_h}$. The weight $\mathbf{m}_t$ of each type (i.e. entity, relation, absolute location and unnecessary word) for word $l_t$ is defined as follows.
\begin{equation}
    \mathbf{m}_t = \text{softmax}( \mathbf{W}_{l1}\sigma(\mathbf{W}_{l0}\mathbf{h}^l_t + \mathbf{b}_{l0}) + \mathbf{b}_{l1}),
\end{equation}
where $\mathbf{W}_{l0} \in \mathbb{R}^{D_{l0} \times D_{h}}$, $\mathbf{b}_{l0} \in \mathbb{R}^{D_{l0} \times 1}$, $\mathbf{W}_{l1} \in \mathbb{R}^{4 \times D_{l0}}$ and $\mathbf{b}_{l1} \in \mathbb{R}^{4 \times 1}$ are learnable parameters, $D_{l0}$ and $D_h$ are hyper-parameters and $\sigma$ is the activation function. The feature vector of a phrase is computed as the mean embedding feature (context) of words appearing in the phrase. The set of features for all phrases in the expression is denoted as $\mathbf{F}^q = \{\mathbf{f}^q_m\}_{m=1}^M$ (contextual embeddings $\mathbf{H}^q = \{\mathbf{h}^q_m \in \mathbb{R}^{D_h}\}_{m=1}^M$).

Next, CMRIN computes the language context of every vertex in graph $G^s$ from both words and phrases. When words are considered, on the basis of the word embedding $\mathbf{F}^l = \{\mathbf{f}^l_t\}_{t=1}^T$ and the entity weights of words $\{\mathbf{m}_t^{(0)}\}_{t=1}^T$, a weighted normalized attention distribution over the vertices of graph $G^s$ is defined as follows.
\begin{equation}
\begin{aligned}
    \alpha^{l}_{t,i} &= \mathbf{W}^{l}_n[\text{tanh}(\mathbf{W}^{l}_{v}\mathbf{x}_i^s + \mathbf{W}^{l}_{f}\mathbf{f}^{l}_t)], \\
    \lambda^l_{t,i} &= \mathbf{m}_t^{(0)}
                           \frac{\text{exp}(\alpha^l_{t,i})}
                           {\sum_i^K {\text{exp}(\alpha^l_{t,i})}},
\end{aligned}
\end{equation}
where $\mathbf{W}^l_{n} \in \mathbb{R}^{1 \times D_{n}}$, $\mathbf{W}^l_{v} \in \mathbb{R}^{D_{n} \times D_{x}}$ and $\mathbf{W}^l_{f} \in \mathbb{R}^{D_{n} \times D_{h}}$ are transformation matrices and $D_n$ is hyper-parameter. $\lambda_{t,i}$ is the weighted normalized attention, indicating the probability that word $l_t$ refers to vertex $v_i$. Likewise, CMRIN computes $\alpha^{q}_{m,i}$ for the phrases $Q=\{q\}_i^M$ on the basis of their features $\mathbf{F}^q = \{\mathbf{f}^q_m\}_{m=1}^M$, and the normalized distribution over the vertices are computed as follows. 
\begin{equation}
\begin{aligned}
    \lambda^q_{m,i} &=                   \frac{\text{exp}(\alpha^q_{m,i})}
                           {\sum_i^K {\text{exp}(\alpha^q_{m,i})}},
\end{aligned}
\end{equation}
The language context $\mathbf{c}_i$ at vertex $v_i$ is computed by aggregating all attention weighted word contexts and phrase contexts.
\begin{equation}
    \mathbf{h}_i = \frac{\sum_{t=1}^{T}\lambda^l_{t,i}\mathbf{h}^l_t + \sum_{m=1}^{M}\lambda^q_{m,i}\mathbf{h}^q_m}{\sum_{t=1}^{T}\lambda^l_{t,i}+\sum_{m=1}^{M}\lambda^q_{m,i}}
\end{equation}

Then, the global language context $\mathbf{h}_g$ of graph $G^s$ is calculated as follows.
\begin{equation}
    \mathbf{h}_g = \sum_{t=0}^{T}(\mathbf{m}_t^{(0)} + \mathbf{m}_t^{(1)} + \mathbf{m}_t^{(2)})\mathbf{h}_t^{l}
\end{equation}
where the entity weight, relation weight and absolute location weight are the first three elements of $\mathbf{m}_t$. CMRIN computes the global context only from word contexts because phrases are only used for improving the accuracy of vertex highlighting in the relation graphs.

\subsubsection{Language-Guided Visual Relation Graph}\label{sec:lgvrg}
Different object proposals and different relationships between proposals do not have equal contributions in solving grounding referring expressions. The proposals and relationships mentioned in the referring expression should be given more attention. Our CMRE highlights the vertices and edges of the spatial relation graph $G^s$, that have connections with the referring expression, to generate the language-guided visual relation graph $G^v$. The highlighting operation is implemented by designing a gate for each vertex and edge in graph $G^s$.

The gate $p^{v}_i$ for vertex $v_i$ is defined as the sum over the weighted probabilities that individual words and phrases in the expression refer to vertex $v_i$,
\begin{equation}
    p^{v}_i = \sum_{t=1}^{T}\lambda^{l}_{t,i} + \sum_{m=1}^{M}\lambda^{q}_{m,i}
\end{equation}

Each edge has its own type and the gates for edges are formulated as the gates for edges' types. The weighted normalized distribution of words over the edges of graph $G^s$ is defined as follows.
\begin{equation}
   \mathbf{w}^{e}_{t} = \text{softmax}( \mathbf{W}_{e1}\sigma(\mathbf{W}_{e0}\mathbf{h}^{l}_t + \mathbf{b}_{e0}) + \mathbf{b}_{e1})\mathbf{m}_t^{(1)},
\end{equation}
where $\mathbf{W}_{e0} \in \mathbb{R}^{D_{e0} \times D_{h}}$, $\mathbf{b}_{e0} \in \mathbb{R}^{D_{e0} \times 1}$, $\mathbf{W}_{e1} \in \mathbb{R}^{N_e \times D_{e0}}$ and  $\mathbf{b}_{e1} \in \mathbb{R}^{N_e \times 1}$ are learnable parameters, and $D_{e0}$ is hyper-parameter. $w_{t,j}^{e}$ is the $j$-th element of ${\mathbf{w}^{e}_t}$, which is the weighted probability of word $l_t$ referring to edge type $j$. And the gate $p^{e}_j$ for edges with type $j \in \{1,2,..N^e\}$ is the sum over all the weighted probabilities that individual words in the expression refer to edge type $j$,
\begin{equation}
   p^{e}_j = \sum_{t=1}^{T}w_{t,j}^{e}.
\end{equation}

The language-guided visual relation graph is defined as $G^v = (V, E, \mathbf{X}, P^{v}, P^{e})$, where $P^{v} = \{p^{v}_i\}_{i=1}^K$,  and $P^{e} = \{p^{e}_j\}_{j=1}^{N_e}$.

\subsection{Multimodal Context Modeling}
Our proposed Gated Graph Convolutional Network (GGCN) further fuses the language context into the language-guided visual relation graph to generate multimodal relation graph $G^m$, and computes a multimodal semantic context for every vertex by performing gated graph convolutional operations on the graph $G^m$.

\subsubsection{Language-Vision Feature}
As suggested by visual relationships detection~\cite{dai2017detecting, zellers2018neural}, the spatial locations together with the appearance features of objects are the key indicators of visual relationship, and the categories of objects is highly predictive of relationship. Our GGCN fuses the language context of vertices into the language-guided visual relation graph $G^v$ ($G^v$ encodes the spatial relationships and appearance features of proposals) to generate multimodal relation graph $G^m$, which forms the basis for computing the semantic context of vertices.

We define feature $\mathbf{x}_i^m$ at vertex $v_i$ in $G^m$ to be the concatenation of the visual feature $\mathbf{x}_i^s$ at vertex $v_i$ in the language-guided visual relation graph and the language context $\mathbf{h}_i$ at vertex $v_i$, i.e. $\mathbf{x}_i^m = [\mathbf{x}_i^s, \mathbf{h}_i]$. The multimodal graph is defined as $G^m = (V, E, \mathbf{X}^m, P^{v}, P^{e})$, where $\mathbf{X}^m = \{\mathbf{x}^m_i\}_{i=1}^{K}$.

\subsubsection{Semantic Context Modeling}\label{sec:scm}
Multi-order relationships may exist in referring expressions. We obtain semantic context representing multi-order relationships through message passing. On one hand, semantic features are obtained by learning to fuse the spatial relations, visual features and language features. On the other hand, context representing multi-order relationships is computed by propagating pairwise context in graph $G^m$.

Inspired by Graph Convolutional Network (GCN) for classification~\cite{kipf2017semi-supervised, wang2018zero}, our GGCN adopts graph convolutional operations in multimodal relation graph $G^m$ for computing semantic context. Different from GCN operating in unweighted graphs, GGCN operates in weighted directed graphs with extra gate operations. The $n$-th gated graph convolution operation at vertex $v_i$ in graph $G^m = (V, E, \mathbf{X}^m, P^{v}, P^{e})$ is defined as follows.
\begin{equation}
\begin{aligned}
    \overrightarrow{\mathbf{x}_i}^{(n)} &= \sum_{e_{i,j} > 0} {p^{e}_{e_{i,j}}} (\overrightarrow{\mathbf{W}}^{(n)}\hat{\mathbf{x}}_j^{(n-1)} p^{v}_j + \mathbf{b}^{(n)}_{e_{i,j}}), \\
    \overleftarrow{\mathbf{x}_i}^{(n)} &= \sum_{e_{j,i} > 0} {p^{e}_{e_{j,i}}} (\overleftarrow{\mathbf{W}}^{(n)}\hat{\mathbf{x}}_j^{(n-1)} p^{v}_j + \mathbf{b}^{(n)}_{e_{j,i}}), \\
    \Tilde{\mathbf{x}}_i^{(n)} &= \widetilde{\mathbf{W}}^{(n)}\hat{\mathbf{x}}^{(n-1)}_i + \widetilde{\mathbf{b}}^{(n)}, \\
    \hat{\mathbf{x}}^{(n)}_i & = \sigma(\overrightarrow{\mathbf{x}_i}^{(n)} + \overleftarrow{\mathbf{x}_i}^{(n)} + \Tilde{\mathbf{x}}_i^{(n)}),
\end{aligned}
\end{equation}
where $\hat{\mathbf{x}}^{(0)}_i = \mathbf{x}_i^m$, $\overrightarrow{\mathbf{W}}^{(n)}, \overleftarrow{\mathbf{W}}^{(n)}, \widetilde{\mathbf{W}}^{(n)} \in \mathbb{R}^{D_e \times (D_x + D_h)}$ $\{\mathbf{b}_j^{(n)}\}_{j=1}^{N_e}, \widetilde{\mathbf{b}}^{(n)} \in \mathbb{R}^{D_e \times 1}$ are learnable parameters, and $D_e$ is hyper-parameter. $\overrightarrow{\mathbf{x}_i}^{(n)}$ and $\overleftarrow{\mathbf{x}_i}^{(n)}$ are encoded features for out- and in- relationships respectively. $\Tilde{\mathbf{x}}_i^{(n)}$ is the updated feature for itself. The final encoded feature $\hat{\mathbf{x}}_i^{(n)}$ is the sum of the above three features and $\sigma$ is the activation function. By performing the gated graph convolution operation multiple iterations ($N$), semantic context representing multi-order relationships among vertices can be computed. Such semantic context are denoted as $\mathbf{X}^c = \{\mathbf{x}^{c}_{i} = \hat{\mathbf{x}}^{(N)}_i\}_{i=1}^K$.

Finally, for each vertex $v_i$, we concatenate its encoded spatial feature $\mathbf{p}_i$ mentioned before and its language-guided semantic context $\mathbf{x}_i^c$ to obtain the multimodal context $\mathbf{x}_i = [\mathbf{W}_p\mathbf{p}_i, \mathbf{x}_i^c]$, where $\mathbf{W}_p \in \mathbb{R}^{D_p \times 5}$ and $D_p$ is hyper-parameter.

\subsection{Loss Function}
The matching score between proposal $o_i$ and expression $L$ is defined as follows,
\begin{equation}
    s_i = \text{L2Norm}(\mathbf{W}_{s0}\mathbf{x}_i) \odot \text{L2Norm}(\mathbf{W}_{s1}\mathbf{h}_g),
\end{equation}
where $\mathbf{W}_{s0} \in \mathbb{R}^{D_s \times (D_p + D_x)}$ and $\mathbf{W}_{s0} \in \mathbb{R}^{D_s \times D_h}$ are transformation matrices, and $D_s$ is hyper-parameter.

Inspired by the deep metric learning algorithm for face recognition in \cite{schroff2015facenet}, we adopt the triplet loss with online hard negative mining to train our CMRIN model. The triplet loss is defined as
\begin{equation}
    loss = \text{max}(s_{neg} + \Delta - s_{gt}, 0),
\end{equation}
where $s_{gt}$ and $s_{neg}$ are the matching scores of the ground-truth proposal and the negative proposal respectively. The negative proposal is randomly chosen from the set of online hard negative proposals, $\{o_j | s_j + \Delta - s_{gt} > 0\}$, where $\Delta$ is the margin. During testing, we predict the target object by choosing the object proposal with the highest matching score.

\section{Experiments}

\begin{table*}
\begin{center}
\resizebox{1.0\textwidth}{!}
{
\begin{tabular}{|c|l|c|c|c|c|c|c|c|c|c|}
\hline
& &  &\multicolumn{3}{|c|}{RefCOCO} & \multicolumn{3}{|c|}{RefCOCO+} & \multicolumn{2}{|c|}{RefCOCOg}  \\ \hline
& & feature & val & testA & testB & val & testA & testB & val & test \\ \hline
1 & MMI \cite{mao2016generation} & vgg16 & - & 71.72 & 71.09 & - & 58.42 & 51.23 & - & - \\
2 & Neg Bag \cite{nagaraja2016modeling} & vgg16 & 76.90 & 75.60 & 78.00 & - & - & - & - & 68.40 \\
3 & CG \cite{luo2017comprehension} & vgg16 & - & 74.04 & 73.43 & - & 60.26 & 55.03 & - & - \\
4 & Attr \cite{Liu_2017_ICCV} & vgg16 & - & 78.85 & 78.07 & - & 61.47 & 57.22 & - & - \\
5 & CMN \cite{hu2017modeling} & vgg16 & - & 75.94 & 79.57 & - & 59.29 & 59.34 & - & - \\
6 & Speaker \cite{yu2016modeling} & vgg16 & 76.18 & 74.39 & 77.30 & 58.94 & 61.29 & 56.24 & - & - \\
7 & Listener \cite{yu2017joint} & vgg16 & 77.48 & 76.58 & 78.94 & 60.50 & 61.39 & 58.11 & 69.93 & 69.03 \\
8 & Speaker+Listener+Reinforcer \cite{yu2017joint} & vgg16 & 79.56 & 78.95 & 80.22 & 62.26 & 64.60 & 59.62 & 71.65 & 71.92  \\
9 & VC \cite{zhang2018grounding}  & vgg16 & - & 78.98 & \textcolor{blue}{82.39} & - & 62.56 & \textcolor{blue}{62.90} & - & -  \\
10 & A-ATT \cite{deng2018visual} & vgg16 & \textcolor{blue}{81.27} & \textcolor{blue}{81.17} & 80.01 & \textcolor{blue}{65.56} & \textcolor{blue}{68.76} & 60.63 & - & - \\
11 & MAttNet \cite{yu2018mattnet}  & vgg16 & 80.94 & 79.99 & 82.30 & 63.07 & 65.04 & 61.77 & \textcolor{blue}{73.04} & \textcolor{blue}{72.79}  \\
12 & Ours CMRIN & vgg16 & \textcolor{red}{83.57} & \textcolor{red}{83.97} & \textcolor{red}{82.69} & \textcolor{red}{71.57} & \textcolor{red}{75.60} & \textcolor{red}{65.56} & \textcolor{red}{75.65} & \textcolor{red}{76.45}  \\ \hline
13 & MAttNet \cite{yu2018mattnet} & resnet101 & 85.65 & 85.26 & 84.57 & 71.01 & 75.13 & 66.17 & 78.10 & 78.12  \\
14 & Ours CMRIN & resnet101 & 86.43 & 87.36 & 85.52 & 75.63 & 80.37 & \textbf{69.58} & 79.72 & 80.85  \\
15 & Ours CMRIN + Semantic Branch & resnet101 & \textbf{86.59} & \textbf{88.17} & \textbf{85.59} & \textbf{76.38} & \textbf{81.44} & 68.81 & \textbf{80.76} & \textbf{81.71} \\ \hline
\end{tabular}
}
\end{center}
\caption{Comparison with the state-of-the-art methods on RefCOCO, RefCOCO+ and RefCOCOg. The two best performing methods using VGG-16 are marked in red and blue. The best performing method using ResNet is marked in bold.}
\label{tab:ref}
\end{table*}

\begin{figure*}[h]
\begin{center}
\includegraphics[width=1.0\linewidth]{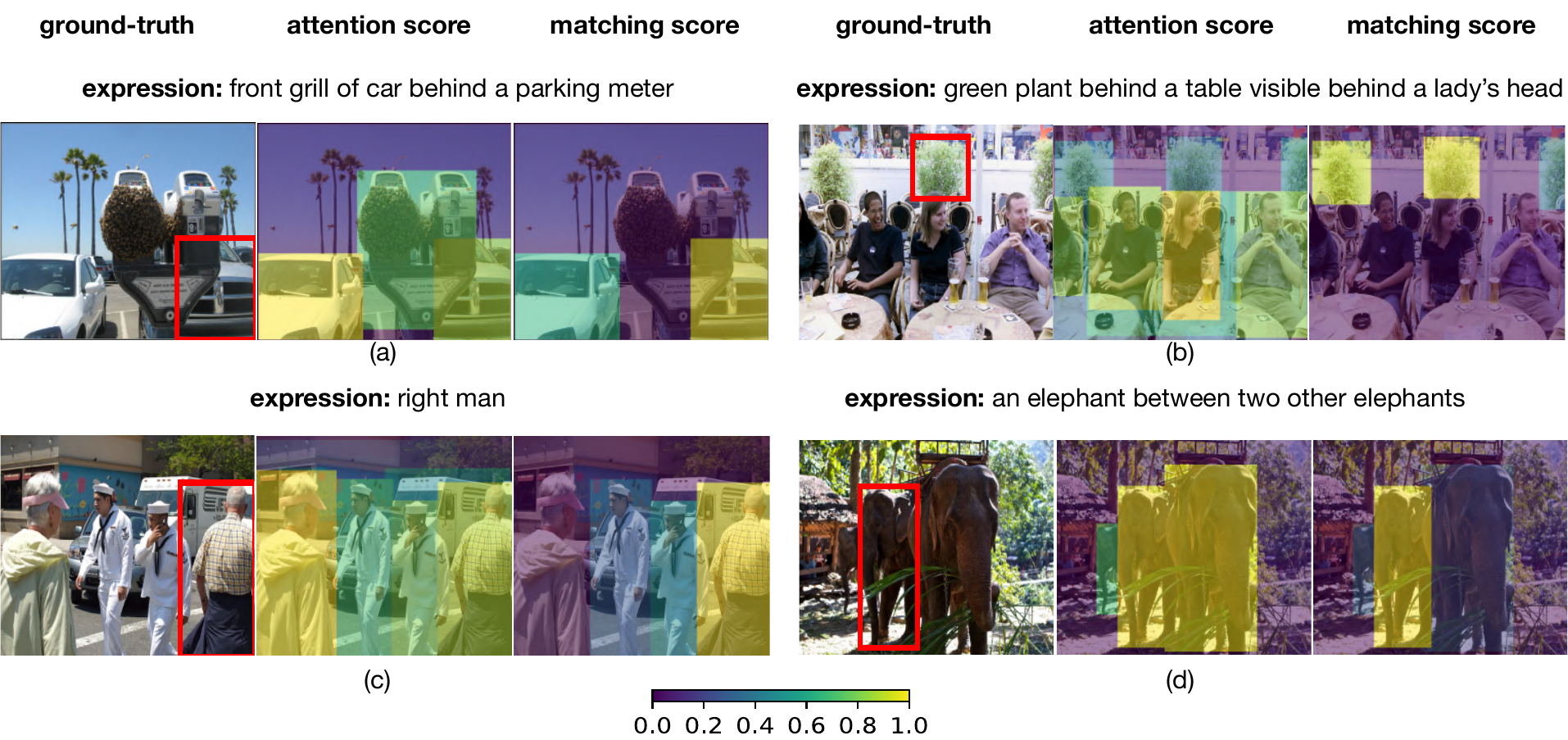}
\end{center}
   \caption{Qualitative results showing initial attention score (gate) maps and final matching score maps. We compute the score of a pixel as the maximum score of proposals covering it, and normalize the score maps to 0 to 1.  Warmer color indicates higher score.}
\label{fig:vis}
\end{figure*}

\begin{table*}
\begin{center}
\resizebox{1.0\textwidth}{!}
{
\begin{tabular}{|c|l|c|c|c|c|c|c|c|c|}
\hline
& &\multicolumn{3}{|c|}{RefCOCO} & \multicolumn{3}{|c|}{RefCOCO+} & \multicolumn{2}{|c|}{RefCOCOg}  \\ \hline
& & val & testA & testB & val & testA & testB & val & test \\ \hline
1 & global langcxt+vis instance & 79.05 & 81.47 & 77.86 & 63.85 & 69.82 & 57.80 & 70.78 & 71.26 \\
2 & global langcxt+global viscxt(2) & 82.61 & 83.22 & 82.36 & 67.75 & 73.21 & 63.06 & 74.29 & 75.23 \\
3 & weighted langcxt+guided viscxt(2) & 86.02 & 86.21 & 84.51 & 73.59 & 78.62 & 68.01 & 77.14 & 78.29 \\
4 & weighted langcxt+guided viscxt(1)+fusion & 85.89 & 87.27 & 84.61 & 74.28 & 79.24 & 69.16 & 79.41 & 79.38 \\
5 & weighted langcxt+guided viscxt(3)+fusion & 86.20 & 87.24 & 84.91 & 75.26 & 80.06 & 69.52 & 79.55 & 80.55 \\
6 & weighted langcxt+guided viscxt(2)+fusion &\textbf{86.43} & \textbf{87.36} & \textbf{85.52} & \textbf{75.63} & \textbf{80.37} & \textbf{69.58} & \textbf{79.72} & \textbf{80.85} \\ \hline
\end{tabular}
}
\end{center}
\caption{Ablation studies on variants of network architecture of our proposed CMRIN on RefCOCO, RefCOCO+ and RefCOCOg. The number following the ``viscxt'' refers to the number of gated graph convolutional layers used in the model.}
\label{tab:ablation}
\end{table*}

\begin{table*}[h]
\begin{center}
\resizebox{1.0\textwidth}{!}
{
\begin{tabular}{|c|l|c|c|c|c|c|c|c|c|}
\hline
& &\multicolumn{3}{|c|}{RefCOCO} & \multicolumn{3}{|c|}{RefCOCO+} & \multicolumn{2}{|c|}{RefCOCOg}  \\ \hline
& & val & testA & testB & val & testA & testB & val & test \\ \hline
1 & softmax loss & 85.43 & 86.81 & 84.47 & 74.23 & 79.74 & 67.78 & 78.94 & 79.37\\
2 & triplet loss (0.2; 1; random hard) & 86.25 & 87.50 & 84.99 & 75.15 & 80.11 & 68.91 & \textcolor{red}{79.96} & 80.49\\
3 & triplet loss (0.5; 1; random hard) & 85.84 & 86.87 & 84.69 & 74.52 & 79.99 & 68.32 & 79.04 & 79.97\\
4 & triplet loss (0.1; 2; random hard) & 86.32 & \textcolor{blue}{87.54} & 84.97 & 75.12 & \textcolor{red}{80.75} & \textcolor{blue}{69.13} & 79.39 & \textcolor{red}{81.30}\\
5 & triplet loss (0.1; 1; hardest) & \textcolor{blue}{86.38} & \textcolor{red}{87.68} & \textcolor{blue}{85.06} & 74.32 & 79.69 & 68.30 & 79.68 & 80.38 \\
6 & triplet loss (0.1; 1; random semi-hard) & 86.35 & 87.36 & 84.87 & \textcolor{blue}{75.16} & \textcolor{blue}{80.39} & 69.07 & 79.41 & 80.55 \\
7 & triplet loss (0.1; 1; random hard) & \textcolor{red}{86.43} & 87.36 & \textcolor{red}{85.52} & \textcolor{red}{75.63} & 80.37 & \textcolor{red}{69.58} & \textcolor{blue}{79.72} & \textcolor{blue}{80.85} \\ \hline
\end{tabular}
}
\end{center}
\caption{Comparison of different schemes for training our proposed CMRIN on RefCOCO, RefCOCO+ and RefCOCOg. The contents in parentheses following the ``triplet loss'' represent the margin value, number of negative proposals and the sampling strategy respectively. The two best performing models are marked in red and blue. CMRIN is robust to different loss settings and consistently outperforms existing state-of-the-art models on all the three benchmark datasets.}
\label{tab:loss}
\end{table*}

\subsection{Datasets}
We have evaluated our CMRIN on three commonly used benchmark datasets for referring expression comprehension (i.e., RefCOCO~\cite{yu2016modeling}, RefCOCO+~\cite{yu2016modeling} and RefCOCOg~\cite{mao2016generation}).

In RefCOCO, there are 50,000 target objects, collected from 19,994 images in MSCOCO~\cite{lin2014microsoft}, and 142,210 referring expressions, collected from an interactive game interface~\cite{kazemzadeh2014referitgame}. RefCOCO is split into train, validation, test A, and test B, which has 120,624, 10,834, 5,657 and 5,095 expression-target pairs, respectively. Test A includes images of multiple people while test B contains images with multiple other objects.

RefCOCO+ has 49,856 target objects collected from 19,992 images in MSCOCO, and 141,564 expressions collected from an interactive game interface. Different from RefCOCO, RefCOCO+ does not contain descriptions of absolute location in the expressions. It is split into train, validation, test A, and test B, which has 120,191, 10,758, 5,726 and 4,889 expression-target pairs, respectively.

RefCOCOg includes 49,822 target objects from 25,799 images in MSCOCO, and 95,010 long referring expressions collected in a non-interactive setting. RefCOCOg~\cite{nagaraja2016modeling} has 80,512, 4,896 and 9,602 expression-target pairs for training, validation, and testing, respectively.

\subsection{Evaluation and Implementation}
The Precision@1 metric (the fraction of correct predictions) is used for measuring the performance of a method for grounding referring expressions. A prediction is considered to be a true positive when the Intersection over Union between the ground-truth proposal and the top predicted proposal for a referring expression is larger than 0.5.

For a given dataset, we count the number of occurrences of each word in the training set. If a word appears more than five times, we add it to the vocabulary. 
Each word in the expression is initially an one-hot vector, which is further converted into a word embedding. We parse the expression into a constituency tree by Stanford CoreNLP toolkit~\cite{manning-EtAl:2014:P14-5}. Annotated regions of object instances are provided in RefCOCO, RefCOCO+ and RefCOCOg. The target objects in the three datasets belong to the 80 object categories in MSCOCO, but the referring expressions may include objects beyond the 80 categories. In order to make the scope of target objects consistent with referring expressions, it is necessary to recognize objects in expressions, even when they are not in the 80 categories.

Inspired by the Bottom-Up Attention Model in \cite{Anderson2017up-down} for image captioning and visual question answering, we train ResNet-101 based Faster R-CNN~\cite{he2016deep, ren2015faster} over selected 1,460 object categories in the Visual Genome dataset \cite{krishna2017visual}, excluding the images in the training, validation and testing sets of RefCOCO, RefCOCO+ and RefCOCOg. 
We combine the detected objects and the ground-truth objects provided by MSCOCO to form the final set of objects in the images. We extract the visual features of objects as the 2,048-dimensional output from the pool5 layer of the ResNet-101 based Faster R-CNN model. Since some previous methods use VGG-16 as the feature extractor, we also extract the 4,096-dimensional output from the fc7 layer of VGG-16 for fair comparison.
We set the mini-batch size to 64. The Adam optimizer~\cite{kingma2014adam} is adopted to update network parameters with the learning rate set to 0.0005 initially and reduced to 0.0001 after 5 epochs. Margin $\Delta$ is set to $0.1$.

\subsection{Comparison with the State of the Art}
We compare the performance of our proposed CMRIN against the state-of-the-art methods, including MMI \cite{mao2016generation}, Neg Bag \cite{nagaraja2016modeling}, CG \cite{luo2017comprehension}, Attr \cite{Liu_2017_ICCV}, CMN \cite{hu2017modeling}, Speaker \cite{yu2016modeling}, Listener \cite{yu2017joint}, VC \cite{zhang2018grounding}, A-ATT \cite{deng2018visual} and MAttNet \cite{yu2018mattnet}.
\subsubsection{Quantitative Evaluation}
Table~\ref{tab:ref} shows quantitative evaluation results on RefCOCO, RefCOCO+ and RefCOCOg datasets. Our proposed CMRIN consistently outperforms existing methods across all the datasets by a large margin, which indicates that our CMRIN performs well on datasets with different characteristics. 
Specifically, CMRIN improves the average Precision@1 over validation and testing sets achieved by the existing best-performing algorithm by 1.80\%, 5.17\% and 3.14\% respectively on the RefCOCO, RefCOCO+ and RefCOCOg datasets when VGG-16 is used as the backbone network. Our CMRIN significantly improves the precision in the person category (test A of RefCOCO and RefCOCO+), which indicates that casting appearance attributes (e.g. shirt, glasses and shoes) of a person as external relationships between person and appearance attributes can effectively distinguish the target person from other persons. After we switch to the visual features extracted by ResNet-101 based Faster R-CNN, the Precision@1 of our CMRIN is further improved by another \textcolor{black}{$\sim$4.40\%}.  It improves the average Precision@1 over validation and testing sets achieved by MAttNet~\cite{yu2018mattnet} by \textcolor{black}{1.62\%, 5.03\% and 3.13\%} respectively on the three datasets. Note that our CMRIN only uses the 2048-dimensional features from pool5 while MattNet uses the feature maps generated from the last convolutional layers of both the third and fourth stages.

\begin{figure*}[h]
\begin{center}
\includegraphics[width=1.0\linewidth]{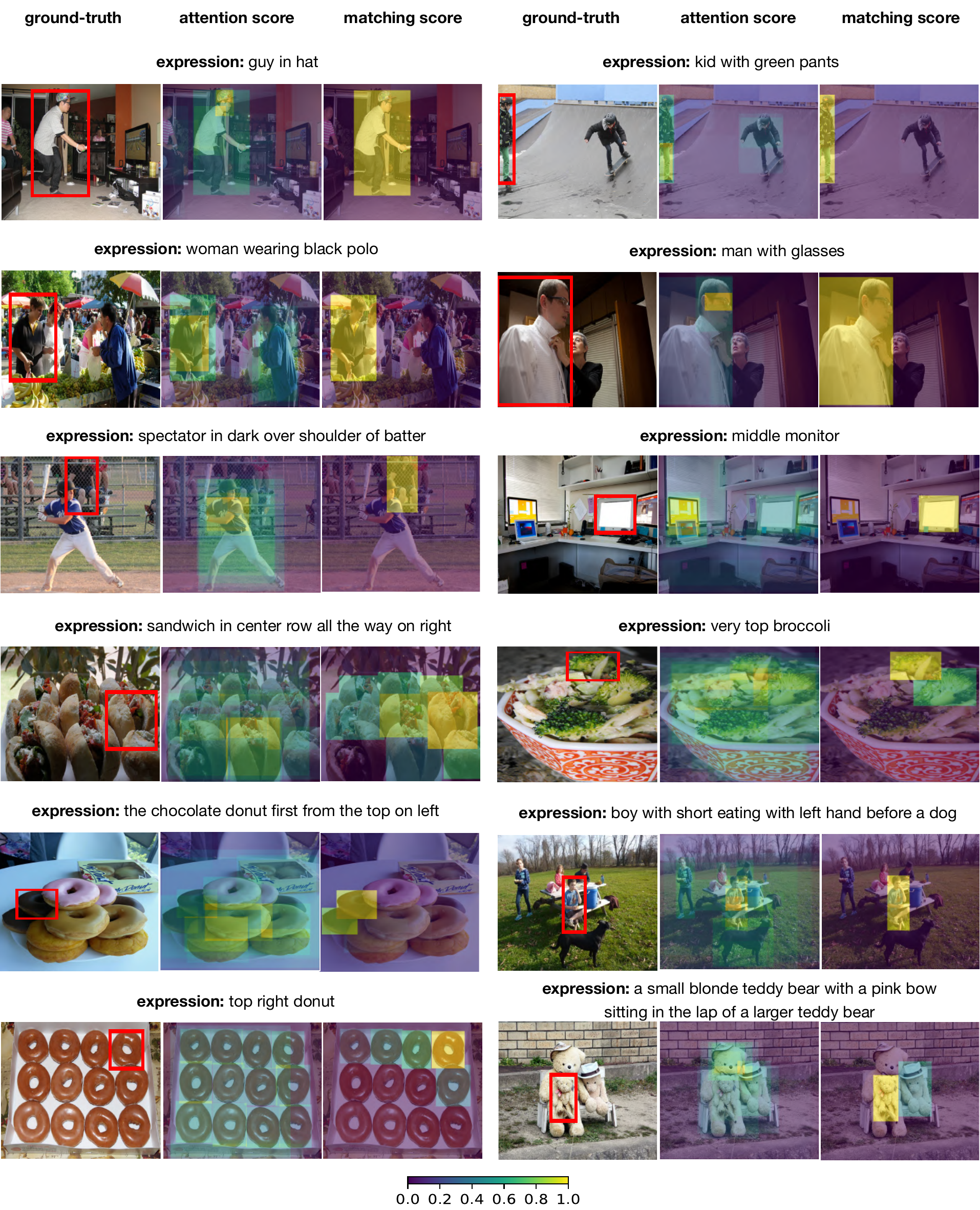}
\end{center}
   \caption{Qualitative results showing initial attention score (gate) maps and final matching score maps. Warmer color indicates higher score.}
\label{fig:more_vis}
\end{figure*}

\subsubsection{Qualitative Evaluation}
Visualizations of some samples along with their attention maps and matching scores are shown in Fig.~\ref{fig:vis}. They are generated from our CMRIN using ResNet-101 based Faster R-CNN features.

Without relationship modeling, our CMRIN can identify the proposals appearing in the given expression~(second columns), and it achieves this goal on the basis of mentioned objects in the given sentence (e.g. the parking meter in Fig.~\ref{fig:vis}(a) and the elephant in full view in Fig.~\ref{fig:vis}(d) have higher attention scores). After fusing information from different modes and propagating multimodal information in the structured relation graph, it is capable of learning semantic context and locating target proposals (third columns) even when the target objects do not attract the most attention at the beginning. It is worth noting that our CMRIN learns semantic relationships (``behind'') for pairs of proposals with different spatial relationships (``bottom right'' between ``car'' and ``parking meter'' in Fig.~\ref{fig:vis}(a); ``top'' between ``green plant'' and ``lady's head'' in Fig.~\ref{fig:vis}(b)), which indicates that CMRIN is able to infer semantic relationships from the initial spatial relationships. In addition, CMRIN learns the context for target ``elephant'' (Fig. \ref{fig:vis}(d)) from ``two other elephants'' by considering the relations from multiple elephants together. Moreover, multi-order relationships are learned through propagation in CMRIN, e.g., the relationships (``right'' in Fig.~\ref{fig:vis}(c)) between object pairs are propagated gradually to the target proposal (most ``right man'').

Fig.~\ref{fig:more_vis} demonstrates more qualitative results from our proposed CMRIN. In order to better visualize the pixels covered by multiple proposals generated from the same object (e.g., the pixels covered by proposal “man” and proposal “shirt weared by the man”), we compute the score of a pixel in attention score maps as the sum of the scores of all covering proposals. And in order to distinguish different objects, we set the score of a pixel in matching score maps to be the maximum among scores of all covering objects. We exclude negative scores and normalize the range of each score map to $[0, 1]$.
%

\subsection{Ablation Study}
\textcolor{black}{We evaluate the proposed CMRIN in five different aspects: 1) the effectiveness of the two modules in our proposed network architecture, i.e., CMRE and GGCN modules; 2) the impact of different training schemes on the performance of CMRIN; 3) the necessity of phrases; 4) the impact of variants of spatial relation graphs used in CMRIN; 5) we explore the effectiveness of incorporating semantic relation graph and detail its implementation. In the following experiments, features computed using ResNet-101 based Faster R-CNN are adopted.}

\subsubsection{Variants of Network Architecture}
Our proposed CMRIN includes CMRE and GGCN modules. To demonstrate the effectiveness and necessity of each module and further compare each module against its variants, we have trained five additional models for comparison. The results are shown in Table \ref{tab:ablation}.

As a baseline (row 1), we use the concatenation of instance-level visual features of objects and the location features as the visual features, and use the last hidden state of the LSTM based expression encoder as the language feature, and then compute the matching scores between the visual features and the language feature. In comparison, a simple variant (row 2) that relies on a global visual context, which is computed by applying graph convolutional operations to the spatial relation graph, already outperforms the baseline. This demonstrates the significance of visual context. Another variant (row 3) with a visual context computed in the language-guided visual relation graph outperforms the above two versions. It captures the context by considering cross-modal information. By fusing the context of words into the language-guided visual relation graph, the semantic context can be captured by applying gated graph convolutional operations (row 6, the final version of CMRIN). Finally, we explore the number of gated graph convolutional layers used in CMRIN. The 1-layer CMRIN (row 4) performs worse than the 2-layer CMRIN because it only captures context with first-order relationships. The 3-layer CMRIN (row 5) does not further improve the performance. One possible reason is that third-order relationships merely occur in the expressions.

\subsubsection{Different Training Schemes}
In this section, we evaluate the impact of different loss function designs on the performance of the proposed CMRIN. As shown in Table~\ref{tab:loss}, CMRIN is robust with respect to different loss function settings~(i.e., the softmax loss and triplet loss with different parameter settings) and consistently outperforms existing state-of-the-art models on all the three commonly used benchmark datasets (i.e. RefCOCO, RefCOCO+ and RefCOCOg). 

Specifically, we compare different loss functions, hyperparameters and sampling strategies in the triplet loss for the training of CMRIN. We optimize the proposed CMRIN with the softmax loss (row 1), which is commonly adopted by existing works~\cite{hu2017modeling, luo2017comprehension, yu2016modeling, zhang2018grounding}. The performance of CMRIN using the softmax loss performs worse than that of CMRIN using the triplet loss (row 7) because the matching score between the context of an expression and the context of a proposal is not always exactly zero or one. For example, in the image associated with expression ``the umbrella held by a lady wearing a green skirt'', there are three umbrellas held by three different ladies and only one of them wears a green skirt. The context of two umbrellas held by the ladies without wearing a green skirt partially matches the context (``the umbrella held by a lady'') of the expression. In addition, we explore the effects of different margins (i.e. 0.1, 0.2 and 0.5) in the triplet loss. CMRIN trained using the triplet loss with a 0.5 margin achieves worse performance  (row 3) than that with other margins (row 2 and 7) over all the three datasets. Moreover, the performance of CMRIN using the triplet loss with a 0.5 margin fluctuates during training. The models trained by the triplet loss with margin 0.1 and 0.2 have similar performance. 
Meanwhile, noting that sampling strategies for the triplet loss are essential in face recognition~\cite{schroff2015facenet,manmatha2017sampling}, we also sample triplets using different online sampling strategies, including random sampling with one hard negative proposal (row 7), random sampling with two hard negative proposals (row 4), hardest negative mining (row 5) and random semi-hard negative mining (row 6; semi-hard negative proposals can be hard and some of their matching scores are smaller than the matching score of the ground-truth proposal). CMRINs using the triplet loss with one or two negative proposals have similar performance. Their differences in average precision over the three testing sets (RefCOCO, RefCOCO+ and RefCOCOg) are -0.19\%, -0.04\% and 0.06\%,  respectively. CMRINs trained using the triplet loss with three different definitions of negative proposals have similar performance except the triplet loss with hardest negative mining on the RefCOCO+ dataset. Their differences in precision over the validation sets and testing sets are within $\pm${0.65\%}, which demonstrates the robustness of our proposed CMRIN with respect to different sampling strategies.

We report the performance of  CMRIN (row 7: triplet loss with a margin of 0.1, one negative proposal and random hard negative mining) as the final version of our algorithm, which is chosen according to its performance on the validation sets.

\subsubsection{Necessity of Phrases}
\begin{table}[h]
	\begin{center}
		\resizebox{0.5\textwidth}{!}
		{
			\begin{tabular}{|l|c|c|c|c|c|}
				\hline
				& \multicolumn{2}{|c|}{RefCOCO} & \multicolumn{2}{|c|}{RefCOCO+} & RefCOCOg  \\ \hline
				& testA & testB & testA & testB  & test \\ \hline
				w/o phrase& 86.95 & 84.40 & 79.57 & 68.23 & 79.71 \\
				implicit phrase & 87.63 & 84.73 & 80.93 & 68.99 & 80.66 \\
				CMRIN & 87.36 & 85.52 & 80.37 & 69.58 & 80.85  \\ \hline
			\end{tabular}
		}
	\end{center}
	\caption{\textcolor{black}{Comparison of different phrase designs on RefCOCO, RefCOCO+ and RefCOCOg.}}
	\label{tab:phrase}
\end{table}

\textcolor{black}{We discuss the necessity of phrases in this section and the results are shown in Table~\ref{tab:phrase}. The performance of the variant using words to highlight the vertices of the spatial relation graph (row 1) is worse than that of the final version using both words and phrases (row 3), which demonstrates the effectiveness of phrases in improving the accuracy of vertex highlighting. It is worth noting that we implicitly considered the word context~(phrase-level information) in our conference version (row 2) by using the contextual features of words to attend the vertices instead of using the word embeddings. However, the contextual features of words introduce the global noise of the expressions, which increases the difficulty of learning the correspondence between words and vertices. The performance of CMRIN with implicit phrases is worse than that of it with explicit phrases in the object category (i.e., test B), because the visual contents of object category is sensitive to contextual noise. In addition, explicit use of phrases can help align between the linguistic words and visual objects.}

\subsubsection{Variants of Spatial Relation Graph}
\begin{table*}[h]
	\begin{center}
		\resizebox{0.9\textwidth}{!}
		{
			\begin{tabular}{|c|l|c|c|c|c|c|c|c|c|}
				\hline
				& &\multicolumn{3}{|c|}{RefCOCO} & \multicolumn{3}{|c|}{RefCOCO+} & \multicolumn{2}{|c|}{RefCOCOg}  \\ \hline
				& & val & testA & testB & val & testA & testB & val & test \\ \hline
				1 & type(7) + center dis(0.5) & 86.50 & 87.56 & 84.26 & 75.18 & 80.58 & 69.01 & 79.58 & 80.37\\
				2 & soft + edge num(5) & \textcolor{red}{86.94} & 88.12 & 84.46 & 75.12 & 80.06 & 68.62 & \textcolor{blue}{80.43} & 80.80 \\
				3 & type(11) + edge num(5) & 86.58 & \textcolor{red}{88.39} & 84.47 & 75.29 & \textcolor{red}{81.63} & 68.64 & 80.11 & \textcolor{blue}{81.07}\\
				4 & type(11) + axis dis(0.15) & 86.46 & 87.91 & \textcolor{blue}{85.14} & \textcolor{red}{76.01} & \textcolor{blue}{81.02} & 68.99 & \textcolor{red}{80.45} & \textcolor{red}{81.21}\\
				5 & type(11) + center dis(0.3) & \textcolor{blue}{86.67} & \textcolor{blue}{88.21} & 84.53 & 75.46 & 80.49 & \textcolor{blue}{69.26} & 80.29 & 80.35\\
				6 & type(11) + center dis(0.7) & 86.57 & 87.71 & 84.14 & 75.21 & 79.74 & 69.16 & 79.68 & 80.20 \\
				7 & type(11) + center dis(0.5) & 86.43 & 87.36 & \textcolor{red}{85.52} & \textcolor{blue}{75.63} & 80.37 & \textcolor{red}{69.58} & 79.72 & 80.85 \\ \hline
			\end{tabular}
		}
	\end{center}
	\caption{\textcolor{black}{Ablation studies on variants of spatial relation graph of our proposed CMRIN on RefCOCO, RefCOCO+ and RefCOCOg. The variant is denoted as design of edge type with condition of existence for edges. The numbers in parentheses following the ``type'', ``dis'' and ``num'' represent the number of types of edges, the threshold of normalized distance and the maximum number of edges at each vertices respectively. The two best performing models are marked in red and blue. CMRIN consistently outperforms existing state-of-the-art models on all the three benchmark datasets.}}
	\label{tab:spatial}
\end{table*}
\textcolor{black}{We conduct experiments for CMRINs with different spatial relation graphs to evaluate the effects of different designs for spatial relation graphs, and those designs come from two perspectives. }

\textcolor{black}{On one hand, we adopt three types of designs for edges, i.e, ``type(11)'', ``type(7)'' and ``soft''. Specifically, ``type(11)'' is the 11 types of edges introduced in Section~\ref{sec:rgc}; the ``type(7)'' is a coarse-grained version of ``type(11)'' and its 7 types of edges are ``inside'', ``cover'', ``overlap'', ``right''', ``top'', ``left'' and ``bottom''; the ``soft'' is a fine-grained version of ``type(11)'' and it directly encodes the edges as relative location representations \cite{yu2018mattnet} by calculating the offsets and area ratios between objects. }
\textcolor{black}{As shown in Table~\ref{tab:spatial}, the performance of CMRIN with ``type(7)'' (row 1) is slightly worse than that of it with ``type(11)'' (row 7), because the design of ``type(7)'' is coarse than the design of ``type(11)''.  The CMRIN with ``soft'' (row 2) and ``type(11)'' (row 3) have similar performance on RefCOCO and RefCOCOg datasets, but the performance of latter is better than that of the former on RefCOCO+ dataset, which indicates that ``type(11)'' is fine enough to capture spatial relationships. In addition, ``type(11)'' is more memory- and computation-efficient than ``soft''.}

\textcolor{black}{On the other hand, we evaluate different conditions for connecting between objects, i.e, ``edge num'', ``axis dis'' and ``center dis''. In particular, the ``edge num(5)'' constraints the maximum out-degrees of each vertices to 5 and a vertex is connected to its 5 nearest nodes based on the distances between their normalized center coordinates (i.e., center distances) \cite{yu2018mattnet}; the ``axis dis(0.15)'' connects each pair of objects as long as the relative distances between them in axes are smaller than 15\% of the length and width of the image respectively \cite{hudson2019learning}; the ``center dis(threshold)'' creates a edge for each pairs of objects whose center distance is smaller than the threshold. As shown in Table~\ref{tab:spatial}, CMRIN with ``center dis(0.7)'' (row 6) has relative lower precision than that with other conditions (row 3, 4, 5 and 7), because the ``center dis(0.7)'' covers several redundant edges which introduces noisy information. The CMRIN with remaining conditions have similar performance, and their differences in average precision over the validation and testing sets on RefCOCO, RefCOCO+ and RefCOCOg datasets are within $\pm${0.07\%}, $\pm${0.27\%} and $\pm${0.55\%}, respectively. }

\subsubsection{Semantic Relation Graph Branch} \label{sec:sg}
\textcolor{black}{It is intuitive to encode the semantic relationships among objects, in this section, we explore the effectiveness and detail the implementation of semantic relation graph branch.}

\begin{table}[h]
	\begin{center}
		\resizebox{0.5\textwidth}{!}
		{
			\begin{tabular}{|l|c|c|c|c|c|}
				\hline
				& \multicolumn{2}{|c|}{RefCOCO} & \multicolumn{2}{|c|}{RefCOCO+} & RefCOCOg  \\ \hline
				& testA & testB & testA & testB  & test \\ \hline
				spatial & 87.36 & 85.52 & 80.37 & 69.58 & 80.85 \\
				semantic & 87.63 & 84.00 & 80.00 & 68.64 & 80.64 \\
				spatial+semantic & 88.17 & 85.59 & 81.44 & 68.81 & 81.71 \\ \hline
			\end{tabular}
		}
	\end{center}
	\caption{\textcolor{black}{Experimental results of spatial/semantic relation graph branch on RefCOCO, RefCOCO+ and RefCOCOg.}}
	\label{tab:semantic}
\end{table}

\textbf{Effectiveness.} \textcolor{black}{We compare the CMRINs with single spatial relation graph branch, with single semantic relation graph branch and joint model including both branches. And the results are shown in the Table~\ref{tab:semantic}. The performance of single semantic branch is worse than that of single spatial branch, because the object relationship detector cannot recognize the semantic relations completely accurately in the highly unrestricted scenes. Moreover, the spatial relation graph branch also implicitly captures the semantic relationships as described in Section~\ref{sec:scm}. In addition, the model including both branches achieves the best precision, which indicate the possibility of cooperation between the spatial and semantic relationship representations.}

\textbf{Implementation.} \textcolor{black}{To implement the branch using semantic relationship graph, we first use a visual relationship detector (the relationship detection part of \cite{yang2018graph}) trained on the Visual Genome datasets excluding the images in RefCOCO, RefCOCO+ and RefCOCOg datasets to extract the semantic relationships among objects. And, for each predicted edge, we compute its features as the probability-weighted embedding of the relationship categories, and the probabilities to relationship categories are predicted by the visual relationship detector. Since we represent each edge as a type in spatial branch and represent each edge as a feature in semantic branch, the implementation for semantic branch has some minor differences with that of spatial branch. In particular, 1) we attend the language representations over the features of edges instead of over the types of edges; 2) We learn the bias vectors for edges by using fully connected layers to encode the edge features in each gated graph convolutional layers instead of learning the bias vectors for types of edges.} 

\textcolor{black}{For model including both branches, the semantic branch and spatial branch share the same language representations and the vertex attention computation, but have their individual attention learning between language representations and visual edges, gated graph convolutional layers and matching computations. The final score of the two-branched model is the mean of matching scores from those two branches. The whole model is end-to-end trained by the triplet loss with online hard negative mining. }

\subsection{Grounding Referring Expressions with Detected Object Proposals}
We have also evaluated the performance of the proposed CMRIN for grounding referring expressions using automatically detected objects in the three datasets. The detected objects are provided by \cite{yu2018mattnet}, and they were detected with a pretrained Faster R-CNN in COCO's training images with the images in the validation and testing sets of RefCOCO, RefCOCO+ and RefCOCOg excluded. The results are shown in Table~\ref{tab:det}. The proposed CMRIN outperforms existing state-of-the-art models, which demonstrates the robustness of CMRIN with respect to object detection results. Specifically, CMRIN improves the average precision in the person category achieved with the existing best-performing method by 3.80\%, and it improves the Precision@1 on RefCOCO+'s test A and test B by 5.50\% and 1.27\%, respectively.

\begin{table}[h]
\begin{center}
\resizebox{0.5\textwidth}{!}
{
\begin{tabular}{|l|c|c|c|c|c|}
\hline
& \multicolumn{2}{|c|}{RefCOCO} & \multicolumn{2}{|c|}{RefCOCO+} & RefCOCOg  \\ \hline
& testA & testB & testA & testB  & test \\ \hline
MMI \cite{mao2016generation} & 64.90 & 54.51 & 54.03 & 42.81 & - \\
Neg Bag \cite{nagaraja2016modeling} & 58.60 & 56.40 & - & - & 49.50 \\
CG \cite{luo2017comprehension} & 67.94 & 55.18 & 57.05 & 43.33 & - \\
Attr \cite{Liu_2017_ICCV} & 72.08 & 57.29 & 57.97 & 46.20 & - \\
CMN \cite{hu2017modeling} & 71.03 & 65.77 & 54.32 & 47.76 & - \\
Speaker \cite{yu2016modeling} & 67.64 & 55.16 & 55.81 & 43.43 & - \\
\textbf{S}+L+R \cite{yu2017joint} & 73.71 & 64.96 & 60.74 & 48.80 & 59.63 \\
VC \cite{zhang2018grounding}& 73.33 & 67.44 & 58.40 & 53.18 & -  \\
MAttNet \cite{yu2018mattnet} & 80.43 & \textbf{69.28} & 70.26 & 56.00 & 67.01  \\
Ours CMRIN & \textbf{82.53} & 68.58  & \textbf{75.76} & \textbf{57.27} & \textbf{67.38}  \\ \hline
\end{tabular}
}
\end{center}
\caption{Comparison with the state-of-the-art methods on RefCOCO, RefCOCO+ and RefCOCOg when using detected proposals. The best performing method is marked in bold.}
\label{tab:det}
\end{table}

\subsection{Effectiveness on multi-order relationships subsets}
To evaluate the effectiveness of our method on multi-order relationships alone, we identify the subset of expressions with indirect references in the RefCOCOg's test set. There are 2,507 expressions with multiple verbs/location words and Precision@1 on this subset is 81.07\% while the number of remaining expression is 7,095 and Precision@1 on them is 80.22\%. In contrast, the Precision@1 of MattNet\cite{yu2018mattnet} (existing best method) is 76.31\% and 79.30\% respectively on these two subsets. This result demonstrates that our method can handle indirect references equally well as other simpler cases.

\section{Conclusions}
In this paper, we focus on the task of referring expression comprehension in images, and demonstrate that a feasible solution for this task needs to not only extract all the necessary information in both the image and referring expressions, but also compute and represent multimodal contexts for the extracted information. In order to overcome the challenges, we propose an end-to-end Cross-Modal Relationship Inference Network (CMRIN), which consists of a Cross-Modal Relationship Extractor (CMRE) and a Gated Graph Convolutional Network (GGCN). CMRE extracts all the required information adaptively for constructing language-guided visual relation graphs with cross-modal attention. GGCN fuses information from different modes and propagates the fused information in the language-guided relation graphs to obtain multi-order semantic contexts. Experimental results on three commonly used benchmark datasets show that our proposed method outperforms all existing state-of-the-art methods.





\ifCLASSOPTIONcaptionsoff
  \newpage
\fi

\bibliographystyle{IEEEtran}
\bibliography{gre_pami}
\vspace{-10mm}
\begin{IEEEbiography}[{\includegraphics[width=1in,height=1.25in,clip,keepaspectratio]{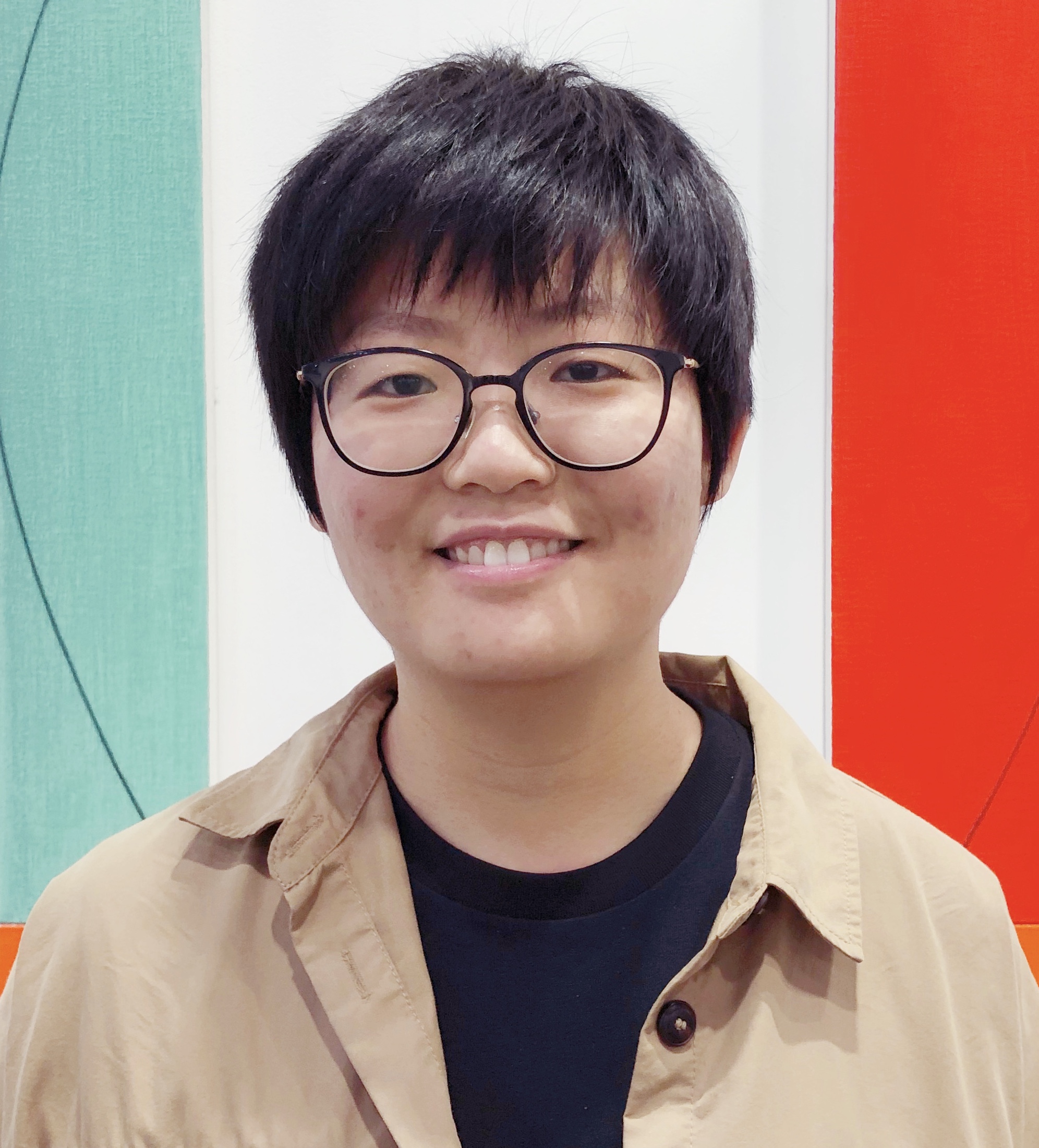}}]{Sibei Yang} received her B.S. degree from School of Computer Science and Technology, Zhejiang University in 2016. She is currently pursuing the Ph.D. degree in the Department of Computer Science, The University of Hong Kong. She is a recipient of Hong Kong PhD Fellowship. She serves as a reviewer for TIP and CVPR. Her current research interests include computer vision, visual and linguistic reasoning, and deep learning. 
\end{IEEEbiography}

\vspace{-10mm}
\begin{IEEEbiography}[{\includegraphics[width=1in,height=1.25in,clip,keepaspectratio]{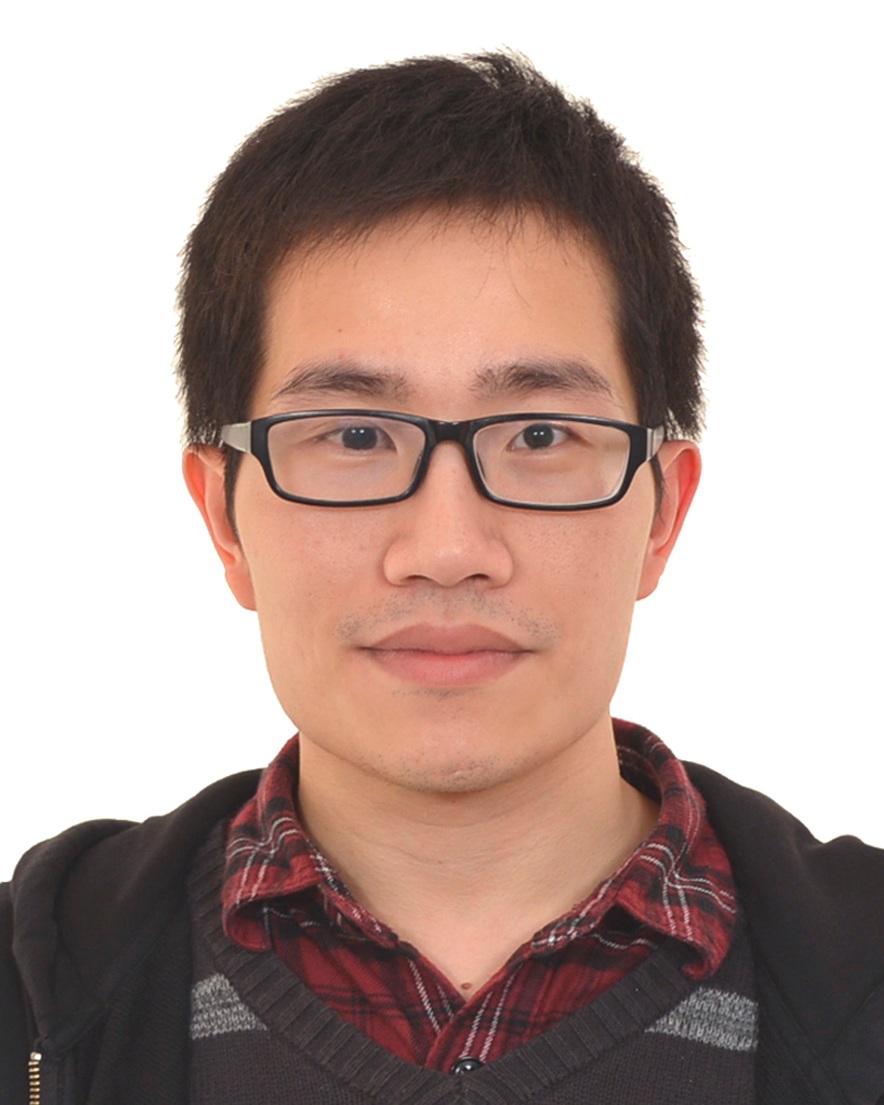}}]{Guanbin Li} (M'15) is currently an associate professor in School of Data and Computer Science, Sun Yat-sen University. He received his PhD degree from the University of Hong Kong in 2016. His current research interests include computer vision, image processing, and deep learning. He is a recipient of ICCV 2019 Best Paper Nomination Award. He has authorized and co-authorized on more than 40 papers in top-tier academic journals and conferences. He serves as an area chair for the conference of VISAPP. He has been serving as a reviewer for numerous academic journals and conferences such as TPAMI, TIP, TMM, TC, CVPR, AAAI and IJCAI.
\end{IEEEbiography}

\vspace{-10mm}
\begin{IEEEbiography}[{\includegraphics[width=1in,height=1.25in,clip,keepaspectratio]{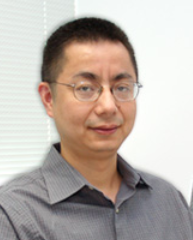}}]{Yizhou Yu} (M'10, SM'12, F'19) received the PhD degree from University of California at Berkeley in 2000. He is a professor at The University of Hong Kong, and was a faculty member at University of Illinois at Urbana-Champaign for twelve years. He is a recipient of 2002 US National Science Foundation CAREER Award, 2007 NNSF China Overseas Distinguished Young Investigator Award, and ACCV 2018 Best Application Paper Award. Prof Yu has served on the editorial board of IET Computer Vision, The Visual Computer, and IEEE Transactions on Visualization and Computer Graphics. He has also served on the program committee of many leading international conferences, including CVPR, ICCV, and SIGGRAPH. His current research interests include computer vision, deep learning, biomedical data analysis, computational visual media and geometric computing.
\end{IEEEbiography}

\end{document}